\documentclass{article}

\usepackage[utf8]{inputenc} 
\usepackage[T1]{fontenc}    
\usepackage{hyperref}       
\usepackage{url}            
\usepackage{booktabs}       
\usepackage{amsfonts}       
\usepackage{nicefrac}       
\usepackage{microtype}      
\usepackage{xcolor}         
\usepackage[numbers]{natbib}
\usepackage{graphicx}
\usepackage{amsmath}
\usepackage{float}
\usepackage[a4paper, total={6in, 8in}]{geometry}
\usepackage{enumitem}
\usepackage[table,xcdraw]{xcolor}
\usepackage[table,xcdraw]{xcolor}
\usepackage{colortbl}
\usepackage[table]{xcolor} 
\usepackage{adjustbox}
\usepackage{bm}

\title{OOD Generalization using Confidence Aware Multi-Teacher Knowledge Distillation}
\date{November 25, 2024}

\author{%
  \textbf{Abdullah Hashmat} \\
  \textit{Research Project}\\
  \texttt{abdullah.hashmat02@gmail.com} \\
  \text{\href{https://github.com/Hashmat02/Multi-Teacher-Single-Student-Confidence-Aware-Knowledge-Distillation}{GitHub Repository Link}}
}

\begin{document}

\maketitle

\begin{abstract}
Our project explores the challenge of OOD generalization in student models for two dominant cues -\textit{texture and shape}- which often limit generalization in cue-conflict scenarios or under distributional shifts during image classification. We propose a multi-teacher single-student knowledge distillation framework where each teacher model is explicitly biased towards a specific visual cue. By combining the strengths of diverse architectures, such as Convolutional Neural Networks (CNNs) for texture, Vision Transformers (ViTs) for shape, and other models for additional biases, the student model learns to integrate and balance these biases. The framework incorporates weighted ensembling of multiple teacher models and intermediate level KD, enabling effective knowledge transfer. The goal is to train a compact and efficient student model that balances these competing biases, rather than inheriting the dominant bias of a single teacher, achieving more consistent generalization across cue-conflict and out-of-distribution variants than any individual teacher or a naively-averaged multi-teacher baseline. This work aims to provide new insights into reconciling architectural diversity and bias-aware training in Machine Learning.
\end{abstract}

\section{Introduction}

\begin{center}
    \textit{"The more you know, the more you know you don't know."} \\
\end{center}

This quote from Aristotle perfectly encapsulates the essence of what we’re tackling today. We strive to build complex models that learn from massive datasets, absorbing layers of knowledge. But as these models get smarter, they reveal gaps and unexpected behaviors when faced with unfamiliar data. This is where we realize that the more we train, the more we uncover how much there is left to learn, especially when tackling real-world challenges like out-of-distribution (OOD) generalization. \\

This can be accounted to the fact that these models often exhibit strong inductive biases towards visual cues such as color, texture, or shape. While these biases are integral for Machine Learning tasks and improve performance on certain datasets, they can limit generalization when data distributions shift or cues conflict. To address these challenges, we explore the potential of a \textbf{multi-teacher Knowledge Distillation} framework. Knowledge Distillation (KD) has emerged as a powerful paradigm for transferring knowledge from one or more high-capacity teacher models to a compact student model, enabling the student to learn distilled insights from the teachers' expertise. Using multiple teacher models, each with distinct inductive biases, the student can synthesize and generalize knowledge across diverse perspectives. This integration of diverse biases equips the student model to handle OOD scenarios more effectively, bridging the gap between theoretical advances in KD and practical demands for robustness in real-world applications. \\

Our project stands at the crossroads of two critical challenges in Machine Learning: OOD generalization and model compression. While the former focuses on enabling models to perform reliably under shifting data distributions, the latter seeks to distill the knowledge of large, complex models into smaller, more efficient ones without sacrificing performance. By combining these domains, our work explores how KD can serve as a bridge - integrating diverse insights from teacher models to train compact student models that are not only efficient but also resilient to distributional shifts.

\section{Related Work}

The approach of enabling the transfer of knowledge from a large, complex teacher model to a smaller, more efficient student model was initially formalized by Hinton \cite{hinton2015distilling}, who proposed matching the softened logits of a teacher to the outputs of a student model, thus allowing the student to mimic the learned representations of the teacher. Multiple efforts have aimed to make student models more generalizable and better performing than their teacher counterparts. For example, self-distillation \cite{furlanello2018born} allows a model to iteratively teach itself, enhancing generalization even without a larger teacher. Adaptive multi-level distillation \cite{liu2020adaptive} refines knowledge representations across multiple layers during training, allowing students to better approximate, and in some cases surpass, their teachers. Weighted ensembling of auxiliary teaching assistants \cite{prasad2022knowledge} similarly boosts the performance and robustness of the student model. \\

An interesting area of emerging research is the use of \textbf{multiple teachers} to distill specialized knowledge to the student. This paper \cite{you2017learning} demonstrated the potential of aggregating knowledge from multiple teachers through an ensemble of logits, thereby capturing diverse feature representations. Such multi-teacher approaches are particularly valuable when individual teachers encode distinct inductive biases. For example, studies \cite{abnar2020transferring} employing Convolutional Neural Networks (CNNs) as teachers emphasize spatially localized features, while transformers prioritize global contextual understanding. This diversity in learned features creates a compelling case for multi-teacher KD to build a more generalized and robust student model. Similarly, \cite{tan2019multilingual} applies multi-teacher learning to multitask learning where each teacher corresponds to a particular task. More recent work has explored dynamic weighting schemes for multi-teacher ensembles, such as gradient-orientation voting \cite{zhou2024govern} and adaptive multi-mentor strategies \cite{sarode2024classroomkd}, though neither explicitly targets reconciling opposing visual-cue biases the way we do. Conversely, \cite{you2018learning} considers the single-teacher, multi-student setting, highlighting how one teacher's biases propagate differently across diverse student architectures, complementary to our focus on multiple teachers distilling into a single student. \\

Recent works include KD from specialized biased teachers to eliminate racial biases in the student model \cite{caldeira2024mst}, by training teachers on a subset of biased data. However, the limitation lies in the homogeneous nature of teacher models which are used, hence model's inherent systematic biases are bound to be propagated to the student model as well. The lack of diversity among teacher models restricts the effectiveness of the distillation process in achieving a relatively unbiased and equitable student model. This paper \cite{wu2021one} pre-trains multiple language models on a specialized NLP task and distills the information to a more generalized student model, however here as well the teacher and student model architectures are homogeneous and are limited to the transformer models only for NLP related tasks. \cite{abnar2020transferring} explores transferring specialized biases from teachers to students using various model architectures. However, it focuses on architectural inductive biases and does not address a multi-teacher setting. This limits its ability to explore how diverse teacher biases can be aggregated to create a more generalized student model. The shape-texture bias gap itself has been studied extensively as a diagnostic property of vision models \cite{geirhos2018imagenet, tartaglini2022developmentally}, but largely as something to measure rather than something to correct via distillation, a gap our work addresses directly. \\

Addressing these gaps, our work seeks to refine the multi-teacher KD paradigm by explicitly leveraging the unique inductive biases of each teacher model while ensuring that these biases complement rather than conflict. By incorporating a dynamic ensemble weighting mechanism and distillation through logit matching \cite{hinton2015distilling} on a cue-conflict dataset \cite{texture-vs-shape}, we aim to produce a student model that not only excels on IID data but also demonstrates better performance on OOD variations.

\section{Methodology}
Since, our work primarily focuses on addressing the challenge of OOD generalization during KD, we target two predominant cues in images: \textbf{texture and shape}. To tackle the issue of student model exhibiting strong biases of teacher models during KD, hindering its ability to generalize effectively in scenarios involving cue conflicts or distributional shifts, we employ \textbf{VGG16}, a CNN model with a known bias toward texture \cite{hermann2020origins}, and \textbf{ViT b16}, a vision transformer, which exhibits a bias toward shape \cite{naseer2021intriguing}, as teacher models. \\

In our framework, the student model learns to balance and integrate these distinct biases through \textbf{confidence-aware weighted logit matching}. The weighting is then dynamically determined based on the entropy levels of the teacher logits \cite{kwon2020adaptive}, enabling the student to adaptively prioritize information from the most confident teacher for each sample. This strategy ensures that the student model develops generalized representations that are less dependent on any single cue, resulting in improved OOD generalization.

\subsection{Our Framework}
In this section, we provide the motivation of learning teacher-dependent importance weight and an overview of the experimental framework.

\subsubsection{Motivation and Overview}
Imagine a classroom where one teacher is an expert in identifying textures, like the complex details of a fabric or the roughness of an object's surface. This teacher excels at recognizing local patterns and fine grained details but struggles to understand the overall structure or shape of objects. Meanwhile, another teacher specializes in recognizing shapes and is good at identifying the global outline of objects like the silhouette of a bird or the shape of a leaf. However, this teacher is less proficient at distinguishing fine textures and surface details between objects.

Now, consider training a student under the guidance of these two teachers. The goal is not to inherit the limitations of either teacher but instead to combine their strengths, so the student can be good at both texture recognition and shape identification. The student should perform well in scenarios where textures are ambiguous, shapes are distorted, or even when the data is OOD. On the whole, we aim to create a student who can efficiently handle both texture and shape based challenges, surpassing the capabilities of either teacher alone.

\begin{figure}[H]
    \centering
    \includegraphics[width=\textwidth]{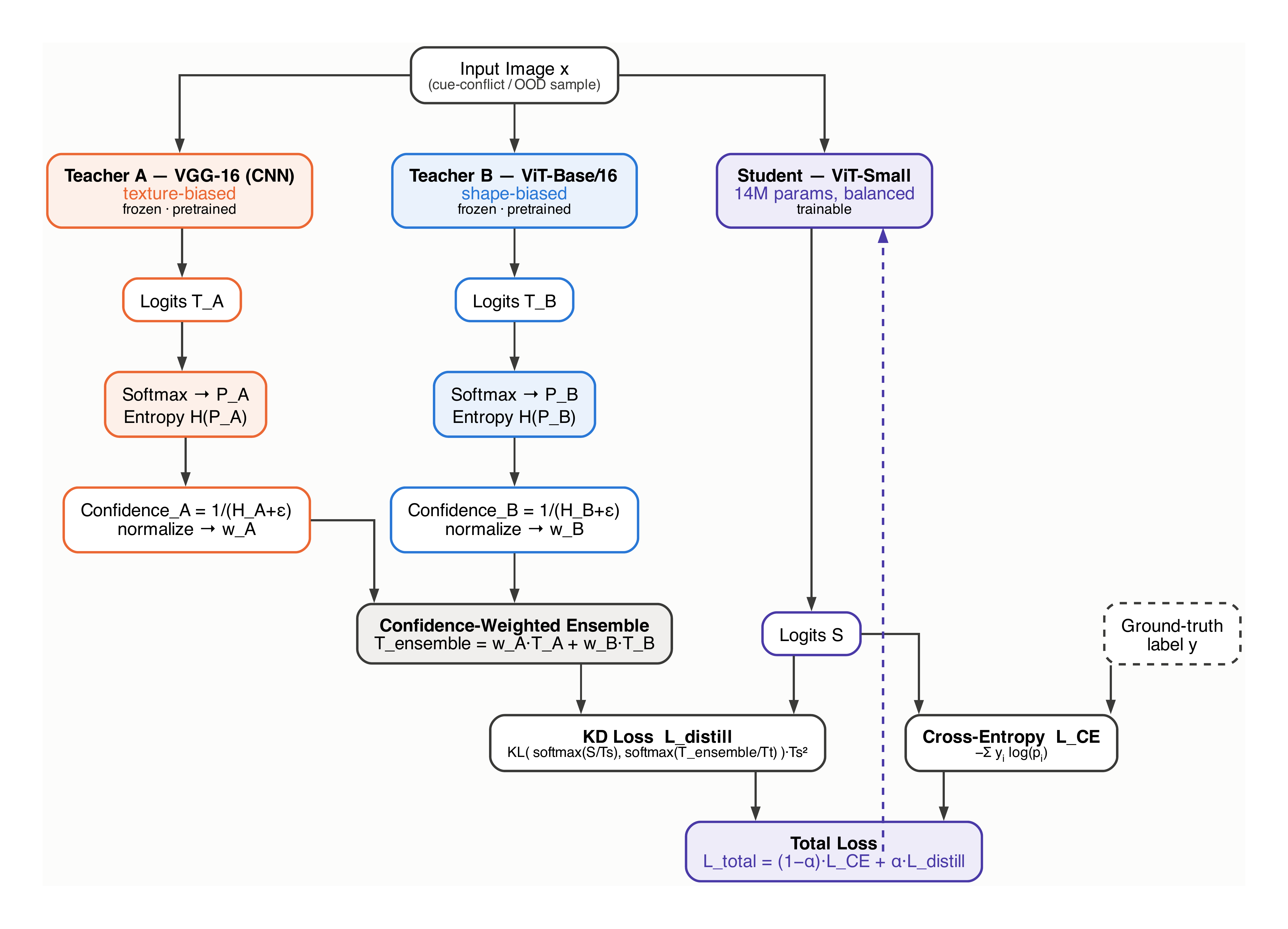}
    \caption{System overview of the confidence-aware multi-teacher KD framework. Both teachers (VGG-16, texture-biased, and ViT-Base/16, shape-biased) are frozen and see the same input. Each teacher's logits are converted to a per-sample confidence score via inverse entropy, normalized into weights, and combined into a single ensembled teacher logit. The student (ViT-Small) is trained against both the ground-truth label (cross-entropy) and the ensembled teacher logit (temperature-scaled KL divergence); the dashed violet arrow denotes backpropagation from the total loss, which updates only the student's parameters.}
    \label{fig:architecture}
\end{figure}

Our goal is to overcome these limitations by leveraging the complementary strengths of shape and texture biased models through a novel knowledge distillation framework. We use two teacher \textbf{models—ViT (shape)} and \textbf{CNN (texture)}—to transfer their biases to a student model via weighted, confidence-aware logit matching. Unlike traditional distillation, where the student inherits the biases of the teachers, our approach balances these biases, creating a more robust student capable of generalizing to \textbf{OOD data} and handling \textbf{cue-conflict} scenarios. Figure~\ref{fig:architecture} summarizes the full pipeline end to end.

\subsubsection{Confidence-Aware Weighting and Knowledge Distillation}
 Here we present our approach that combines entropy based confidence-aware weighting and KD to train an OOD generalizable student model.\\

 \textbf{Logit Matching Loss (KD Loss)}

In Knowledge Distillation (KD), the goal is to transfer knowledge from a teacher model to a student model by aligning their logits (the raw outputs before applying the softmax). We modify this approach by using multiple teacher models and ensembled logits.

The logit matching loss between the student model's logits \( \mathbf{S} \) and the ensembled teacher logits \( \mathbf{T} \) is computed using the Kullback-Leibler (KL) divergence \cite{hinton2015distilling} between the temperature-scaled softmax outputs:

\begin{equation}
L_{\text{distill}} = \text{KL}\left( \text{softmax}\left(\frac{\mathbf{S}}{T_s}\right), \text{softmax}\left(\frac{\mathbf{T}}{T_t}\right) \right) \times T_s^2
\label{eq:distill}
\end{equation}

Where:
\begin{itemize} [itemsep=0.5pt, topsep=1pt]
    \item \( \mathbf{S} \) is the student model's logits.
    \item \( \mathbf{T} \) is the ensembled logits from the teacher models.
    \item \( T_s \) is the temperature factor for the student model.
    \item \( T_t \) is the temperature factor for the teacher models.
    \item The KL divergence measures the difference between the two probability distributions.
    \item \( T_s^2 \) scaling factor is used to adjust for the softened logits.\\
\end{itemize}

This loss encourages the student model to match the softmax outputs of the ensembled teacher models, with temperature scaling making the knowledge transfer smoother.\\

\textbf{Confidence-Aware Weighted Logit Ensemble}

We use an ensemble of multiple teacher models to create the final teacher logits for distillation. Instead of simply averaging the logits, we weight each teacher's contribution based on their confidence level. Confidence is derived from the entropy of the teacher's output distribution.\\

\textbf{1. Softmax:}
For each teacher model \( i \), we compute the softmax probabilities \( \mathbf{P}_i \) of the teacher's logits \( \mathbf{T}_i \):

\begin{equation}
\mathbf{P}_i = \text{softmax}(\mathbf{T}_i)
\label{eq:softmax}
\end{equation}

\textbf{2. Entropy:}
For each teacher's prediction \( \mathbf{P}_i \), we compute the entropy \( H(\mathbf{P}_i) \), which measures the uncertainty in the teacher's prediction, where \( \epsilon \) is a small value to avoid division by zero or log of zero:

\begin{equation}
H(\mathbf{P}_i) = - \sum_j \mathbf{P}_i(j) \cdot \log(\mathbf{P}_i(j) + \epsilon)
\label{eq:entropy}
\end{equation}

\textbf{3. Confidence Calculation:}
The confidence for each teacher is the inverse of their entropy. A teacher with lower entropy ie. more confident predictions will have a higher confidence score:

\begin{equation}
\text{Confidence}_i = \frac{1}{H(\mathbf{P}_i) + \epsilon}
\label{eq:confidence}
\end{equation}

\textbf{4. Normalization of Confidence Scores:}
Rather than a simple linear normalization, the confidence scores are standardized (z-scored) across teacher models and converted to weights via a temperature-scaled softmax, which more strongly separates confident from unconfident teachers on a per-sample basis:

\begin{equation}
z_i = \frac{\text{Confidence}_i - \mu_{\text{Confidence}}}{\sigma_{\text{Confidence}} + \epsilon}
\label{eq:zscore}
\end{equation}

\begin{equation}
\mathbf{w}_i = \frac{\exp(\beta \cdot z_i)}{\sum_{k=1}^{n} \exp(\beta \cdot z_k)}, \qquad \beta = 1.5
\label{eq:weight-norm}
\end{equation}

where \( \mu_{\text{Confidence}} \) and \( \sigma_{\text{Confidence}} \) are the mean and standard deviation of the confidence scores across the \( n \) teachers for a given sample, and \( \beta \) is a fixed softmax temperature.

\textbf{5. Ensemble Logits Calculation:}
The final ensembled teacher logits \( \mathbf{T}_{\text{ensemble}} \) are calculated as a weighted sum of the logits from all teacher models:

\begin{equation}
\mathbf{T}_{\text{ensemble}} = \sum_{i=1}^{n} \mathbf{w}_i \cdot \mathbf{T}_i
\label{eq:ensemble}
\end{equation}

where \( \mathbf{w}_i \) is the normalized weight based on each teacher's confidence, and \( \mathbf{T}_i \) is the logits from teacher \( i \).

This approach ensures that the ensembled logits give more weight to teachers that are more confident in their predictions. It contrasts with other methods where confident teachers might be given lower weight to reduce bias, but we instead emphasize the strong predictions from confident teachers in order to transfer their biases to the student.\\

\textbf{Total Loss Calculation}

The total loss function combines the Cross-Entropy (CE) loss and the logit matching distillation loss.

  \textbf{a) Cross-Entropy Loss:}
This loss measures how well the student model's predictions match the ground-truth labels. It is defined as:

\begin{equation}
L_{\text{CE}} = - \sum_i y_i \log(p_i)
\label{eq:ce-loss}
\end{equation}

Where:
\begin{itemize}[noitemsep, topsep=0pt]
    \item \( y_i \) is the ground-truth label, and
    \item \( p_i \) is the predicted probability for class \( i \).\\
\end{itemize}

\textbf{b) Logit Matching Loss:} This is the distillation loss described earlier. \\

The \textbf{total loss} is a weighted sum of the Cross-Entropy loss and the distillation loss:

\begin{equation}
\boldsymbol{L_{\text{total}}} = (1 - \alpha) \cdot \boldsymbol{L_{\text{CE}}} + \alpha \cdot \boldsymbol{L_{\text{distill}}}
\label{eq:total-loss}
\end{equation}

Where:
\begin{itemize}
    \item \( \alpha \) is a hyperparameter that controls the balance between the classification loss and the distillation loss. By adjusting \( \alpha \), we control how much emphasis is placed on learning from the teacher models versus the ground-truth labels.
\end{itemize}

This dynamic weighting allows the student model to learn from both the teacher models' knowledge and the ground-truth labels, ensuring that the student generalizes well to new data.\\

\textbf{Optimization and Training}

During training, the optimizer (AdamW) is used to minimize the total loss. The optimizer's update rule is:

\begin{equation}
\theta_{\text{new}} = \theta_{\text{old}} - \eta \cdot \nabla_\theta L_{\text{total}}
\label{eq:sgd-update}
\end{equation}

Where:
\begin{itemize}[itemsep=0.5pt, topsep=1pt]
    \item \( \theta_{\text{old}} \) and \( \theta_{\text{new}} \) are the parameters of the student model before and after the update,
    \item \( \eta \) is the learning rate, and
    \item \( \nabla_\theta L_{\text{total}} \) is the gradient of the total loss with respect to the student model parameters.
\end{itemize}

The learning rate is adjusted during training using cosine annealing:

\begin{equation}
\eta_{\text{new}} = \eta_{\text{max}} \cdot \frac{2}{1 + \cos\left(\frac{t}{T_{\text{max}}} \pi \right)}
\label{eq:cosine-lr}
\end{equation}

Where:
\begin{itemize} [itemsep=0.5pt, topsep=1pt]
    \item \( \eta_{\text{max}} \) is the maximum learning rate, and
    \item \( T_{\text{max}} \) is the total number of epochs.
\end{itemize}

\section{Experimental Setup}
This section outlines the experimental setup used to evaluate our multi-teacher knowledge distillation
framework.

\subsection{Datasets}
To comprehensively evaluate the performance of our multi-teacher knowledge distillation framework, we utilized the following datasets: \\

\begin{itemize}[noitemsep, topsep=0pt]
    \item \textbf{ImageNet-1k \cite{deng2009imagenet}:} We used ImageNet-1k for its diverse set of textures and shapes to establish a strong baseline for pretraining and fine-tuning. Using pretrained models on ImageNet also eliminated the need to train models from scratch on this vast dataset.

    \item \textbf{Stylized ImageNet\cite{geirhos2018imagenet}:} This dataset was used to create \textbf{cue-conflict}, shape-texture, scenarios and evaluate model biases. Stylized ImageNet enabled us to evaluate the model's bias towards a certain cue.

    \item \textbf{Canny Edge ImageNet:} This variant of ImageNet consists of edge-detected images generated using the Canny edge detection algorithm . It was used to test the OOD generalization of the models by removing texture details and focusing on shape-based features.

    \item \textbf{Animals10 \cite{berg2006animals}:} This dataset was included due to its rich texture and varying shapes. It served as a test bed for evaluating the generalization capabilities of our student model on unseen data with high visual diversity.

    \item \textbf{Stylized Animals10:} This stylized version of Animals10 was used to evaluate model biases and OOD performance in cue-conflict scenarios.
\end{itemize}

\begin{figure}[H]
    \centering
    \includegraphics[width=\textwidth]{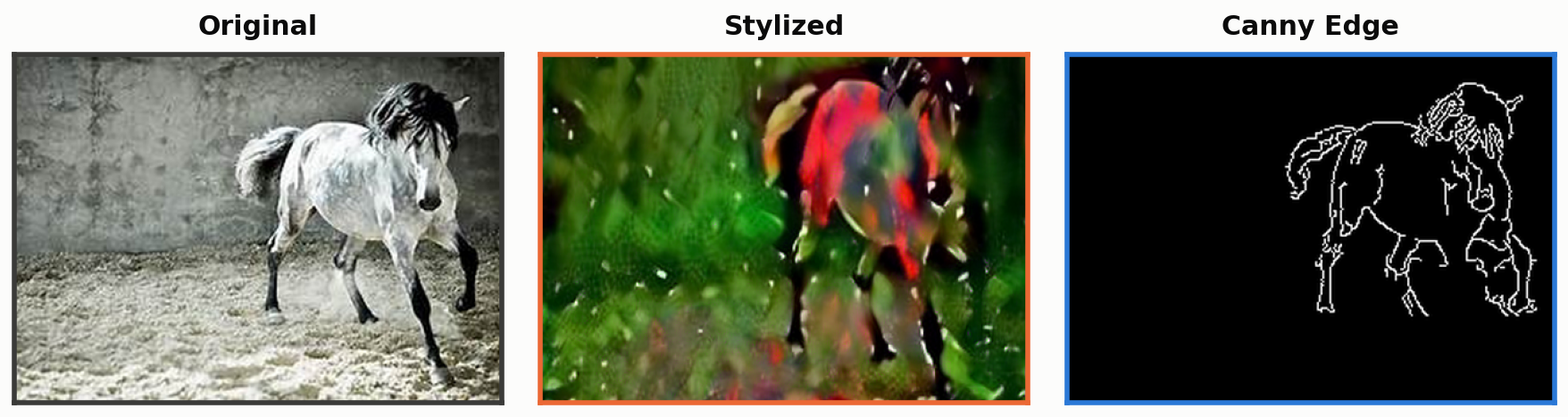}
    \caption{Example of the three evaluation views used for a single Animals-10 sample: the \textbf{original} image, a \textbf{stylized} cue-conflict version produced by our pipeline (content shape from the horse, texture transferred from an unrelated butterfly image), and the corresponding \textbf{Canny edge} map used to isolate shape information.}
    \label{fig:cue-samples}
\end{figure}

\subsection{Models}
In our experiments, we initially explored a variety of architectures to assess their suitability as teacher and student models for our KD framework. The models evaluated included VGG16, VGG11, ResNet50, DenseNet121, Vision Transformer Base Patch (ViT-Base-Patch), and Vision Transformer Small Patch (ViT-Small-Patch). Based on these experiments, we selected the following models:

\begin{itemize}[noitemsep, topsep=0pt]
    \item \textbf{Teachers:} \\
     \textbf{VGG16:} A CNN model, pretrained on ImageNet-1K with a proven bias toward texture features \cite{hermann2020origins}, making it an ideal candidate for transferring texture-specific knowledge.\\
     \textbf{ViT-Base-Patch:} A transformer architecture pretrained on ImageNet-1k, known for its bias toward capturing global shape information \cite{naseer2021intriguing}.
    \item \textbf{Student:} \\
    \textbf{ViT-Small-Patch:} A compact Vision Transformer model with 14M parameters \cite{rw2019timm}. Through experimentation, we observed that this model neither strongly adopted a shape bias nor a texture bias, making it a suitable candidate for learning a balanced representation from the teachers.\\
\end{itemize}

The selection of these models was driven by their complementary inductive biases, which ensured that biases can be propagated from these models. By distilling knowledge from the texture-biased VGG16 and the shape-biased ViT-Base-Patch, we aimed to produce a student model that integrates both biases while remaining compact and efficient for OOD generalization.

\subsection{Baselines}
To evaluate the effectiveness of our proposed multi-teacher KD framework, we compare its performance against several baseline approaches:

\begin{itemize}[noitemsep, topsep=0pt]
    \item \textbf{Single Teacher KD \cite{hinton2015distilling}:}
    We perform KD using a single teacher model at a time. Specifically, the student model is trained using either the texture-biased VGG16 or the shape-biased Vision Transformer (ViT) as the teacher. This approach establishes a baseline for how well the student can generalize when learning from a single source of bias.

    \item \textbf{Multi-Teacher AVG-KD:}
    This baseline involves distilling knowledge from both teacher models using equal-weighted averaging of their logits. By combining the predictions of VGG16 and ViT without considering confidence levels or entropy, this approach provides insight into the limitations of naive ensembling for knowledge transfer.

    \item \textbf{Individual Teacher Model Performance:}
    The performance of the VGG16 and ViT models is evaluated independently on OOD datasets. This serves as a reference for comparing the student model’s ability to generalize beyond the biases inherent to its teachers. These results also highlight the trade-offs between texture and shape biases in individual architectures.
\end{itemize}

\subsection{Evaluation Metrics}
We use the following evaluation metrics to assess the performance of the models and measure the biases on two cues (Shape and Texture):

\begin{itemize}[noitemsep, topsep=0pt]
    \item \textbf{Accuracy:} The accuracy is measured by comparing the model's predictions against the true labels.
    \item \textbf{Cue Accuracy\cite{gavrikov2024vision}:} This is defined as the ratio of predictions that match either the shape or texture label (i.e., correctly classified instances), which is computed as:
    \begin{equation}
    \text{Cue Accuracy} = \text{Shape Accuracy} + \text{Texture Accuracy}
    \label{eq:cue-accuracy}
    \end{equation}
    \item \textbf{Shape Bias:} The shape bias is calculated as the ratio of shape accuracy to cue accuracy:
    \begin{equation}
    \text{Shape Bias} = \frac{\text{Shape Accuracy}}{\text{Cue Accuracy}}
    \label{eq:shape-bias}
    \end{equation}
    \item \textbf{Texture Bias:} The texture bias is simply the complement of the shape bias:
    \begin{equation}
    \text{Texture Bias} = 1 - \text{Shape Bias}
    \label{eq:texture-bias}
    \end{equation}
\end{itemize}

\subsection{Training Procedure and Knowledge Distillation}

In our experiments, we began by fine-tuning both the teacher and student models on the respective datasets, following a standard training procedure. We used \textbf{10 epochs}, a learning rate of \boldmath\( 1 \times 10^{-4} \), and \textbf{batch size} of \textbf{128}. For the Animals10 dataset, both the teacher and student models were trained from scratch, while for ImageNet-1K, we used pretrained models and fine-tuned the classifiers on a subset of 16 superclasses from ImageNet, which were used in the cue-conflict dataset. The subsequent confidence-aware distillation phase (Section~\ref{sec:appendix}) instead used a batch size of 32.

Initially, we measured the accuracies and biases of the models. After training, we observed that VGG16 exhibited a strong texture bias. To induce more shape bias in the Vision Transformer (ViT), we employed curriculum learning as follows:

\begin{itemize} [noitemsep, topsep=0pt]
    \item \textbf{Stage 1:} The model was trained on shape-biased data only.
    \item \textbf{Stage 2:} We mixed shape-biased data with the original data, ensuring that both models developed a shape bias.
\end{itemize}

For the student model, we used ViT-Small, which was initially untrained. The student was fine-tuned for \textbf{2 epochs} on the original data. For the remaining \textbf{8 epochs}, we applied our Confidence-Aware Knowledge Distillation (KD) framework to the student model.

Once training was completed, we measured the biases on the cue-conflict dataset to evaluate the shape and texture biases, as well as the accuracy on the original dataset to assess the impact of OOD generalization.

The experiments were conducted using various setups:
\begin{itemize}[noitemsep, topsep=0pt]
    \item Single Teacher to Student (using VGG16, ViT as the teacher),
    \item Multiple Teachers with Equal Weights to the Student (using VGG16 and ViT),
    \item Base Teachers
    \item Our Method (Confidence aware multi-teacher KD)
\end{itemize}

We compared the biases and accuracy results across these different setups to understand how each method influences the model's ability to generalize, especially in cue-conflict scenarios. Biases were recording across all these setups and accuracies we evaluated on the original, stylized and canny edge variants of \textbf{ImageNet1-k} and \textbf{Animals10}

.

\section{Results and Findings}

Firstly we conducted a series of baseline experiments to investigate the inductive biases of CNN and Vision Transformer (ViT) models. The accuracies for the standard, stylized, and Canny edge datasets were calculated, along with the biases (shape,texture) of the fine-tuned models using the CueConflict datasets respectively. Our results exhibited consistent trends across both datasets we tested it on ie: \textbf{ImageNet-1k} and \textbf{Animals-10}.

\begin{table}[H]
\centering
\begin{adjustbox}{max width=\textwidth}
\begin{tabular}{|>{\columncolor[HTML]{C0C0C0}}l|l|l|l|l|l|l|}
\hline
\textbf{Models} & \multicolumn{3}{|c|}{\cellcolor[HTML]{C4DDFF}{\color[HTML]{333333} \textbf{ImageNet-1K Accuracy}}}     & \multicolumn{3}{|c|}{\cellcolor[HTML]{FFCE93}\textbf{Animals-10 Accuracy}}     \\ \hline
                & \cellcolor[HTML]{DAE8FC}{\color[HTML]{000000} Standard} & \cellcolor[HTML]{DAE8FC}{\color[HTML]{000000} Stylized} & \cellcolor[HTML]{DAE8FC}{\color[HTML]{000000} Canny Edge} & \cellcolor[HTML]{FFE6C8}{\color[HTML]{000000} Standard} & \cellcolor[HTML]{FFE6C8}{\color[HTML]{000000} Stylized} & \cellcolor[HTML]{FFE6C8}{\color[HTML]{000000} Canny Edge} \\ \hline
VGG-16          & \textbf{96.16\%}                                       & \textbf{34.62\%}                                       & 18.33\%                                                & \textbf{73.22\%}                                       & \textbf{27.09\%}                                       & 12.14\%                                                \\ \hline
VGG-11          & 93.42\%                                               & 31.52\%                                               & 12.15\%                                                & 71.41\%                                               & 20.19\%                                               & 11.98\%                                                \\ \hline
ViT-Base Patch  & 86.44\%                                               & 26.02\%                                               & \textbf{55.83\%}                                       & 67.09\%                                               & 12.07\%                                               & \textbf{35.67\%}                                       \\ \hline
ViT-Small Patch & 82.14\%                                               & 22.60\%                                               & 47.29\%                                                & 65.24\%                                               & 9.89\%                                                & 30.12\%                                                \\ \hline
ResNet50        & 86.54\%                                               & 32.11\%                                               & 14.03\%                                                & 70.01\%                                               & 23.44\%                                               & 10.05\%                                                \\ \hline
DenseNet121     & 80.31\%                                               & 28.19\%                                               & 17.22\%                                                & 69.16\%                                               & 19.89\%                                               & 10.63\%                                                \\ \hline
\end{tabular}
\end{adjustbox}
\caption{Base performance comparison of different models on ImageNet-1K and Animals-10 datasets.}
\label{tab:model_comparison}
\end{table}

As seen in Table~\ref{tab:model_comparison}, the larger CNN model, VGG16, outperformed all other models on the standard dataset, achieving an accuracy of \textbf{96.16\%} on ImageNet and \textbf{73.22\%} on Animals-10. This performance can be attributed to the texture-rich nature of both datasets. On the stylized dataset, VGG16 again showed the highest accuracy as expected, achieving \textbf{34.62\%} on ImageNet and \textbf{27.09\%} on Animals-10, due to its ability to capture local features effectively.

On the Canny edge dataset, ViT-Base achieved the highest accuracy, with \textbf{55.83\%} on ImageNet and \textbf{35.67\%} on Animals-10, due to its ability to capture global shape patterns. However, the performance of ViTs on stylized datasets was notably low, with ViT-Small achieving only \textbf{9.89\%} on Animals-10, highlighting its limitations in handling stylized data.

\begin{table}[H]
\centering
\begin{adjustbox}{max width=\textwidth}
\begin{tabular}{|>{\columncolor[HTML]{C0C0C0}}l|>{\columncolor[HTML]{C4DDFF}}c|>{\columncolor[HTML]{C4DDFF}}c|>{\columncolor[HTML]{FFCE93}}c|>{\columncolor[HTML]{FFCE93}}c|}
\hline
\textbf{Models}    & \multicolumn{2}{|c|}{\textbf{ImageNet-1K Bias}}      & \multicolumn{2}{|c|}{\textbf{Animals-10 Bias}}         \\ \hline
                   & {Shape Bias} & {Texture Bias}        & {Shape Bias} & {Texture Bias}         \\ \hline
\textbf{VGG-16}    & \textbf{0.14}       & \textbf{0.86}               & \textbf{0.20}       & \textbf{0.80}                 \\ \hline
\textbf{VGG-11}    & 0.21                & 0.79                        & 0.27                & 0.73                          \\ \hline
\textbf{ViT-Base}  & \textbf{0.64}       & \textbf{0.36}               & \textbf{0.59}       & \textbf{0.41}                 \\ \hline
\textbf{ViT-Small} & 0.59                & 0.41                        & 0.56                & 0.44                          \\ \hline
\textbf{ResNet50}  & 0.29                & 0.71                        & 0.31                & 0.69                          \\ \hline
\textbf{DenseNet}  & 0.23                & 0.77                        & 0.38                & 0.62                          \\ \hline
\end{tabular}
\end{adjustbox}
\caption{Bias (Shape, Texture) comparison for ImageNet-1K and Animals-10 datasets.}
\label{tab:bias_comparison}
\end{table}

In terms of model biases in Table~\ref{tab:bias_comparison}, we observed consistent results across both datasets as well. The VGG16 model, pretrained on ImageNet, exhibited a strong texture bias of \textbf{86\%}, with a minimal shape bias of \textbf{14\%}. In contrast, the ViT-Base model demonstrated a much more balanced bias, with \textbf{64\%} shape bias and \textbf{36\%} texture bias. This trend was similarly reflected in the Animals-10 dataset, where CNNs exhibited strong texture bias, while ViTs were shape-biased.

Overall, our findings confirmed that CNN models, such as VGG16, are predominantly biased towards texture, whereas Vision Transformers like ViT-Base exhibit a greater reliance on shape features, providing a strong baseline for our distillation results. To our surprise, ViT-Small Patch exhibited a relatively balanced bias compared to all other models, with \textbf{59\%} shape bias and \textbf{41\%} texture bias on ImageNet, and \textbf{56\%} shape bias and \textbf{44\%} texture bias on Animals-10. However, the theme of being shape-biased is persistent here as well. These results set up a strong baseline for testing our proposed methodology.

\begin{figure}[H]
    \centering
    \includegraphics[width=\textwidth]{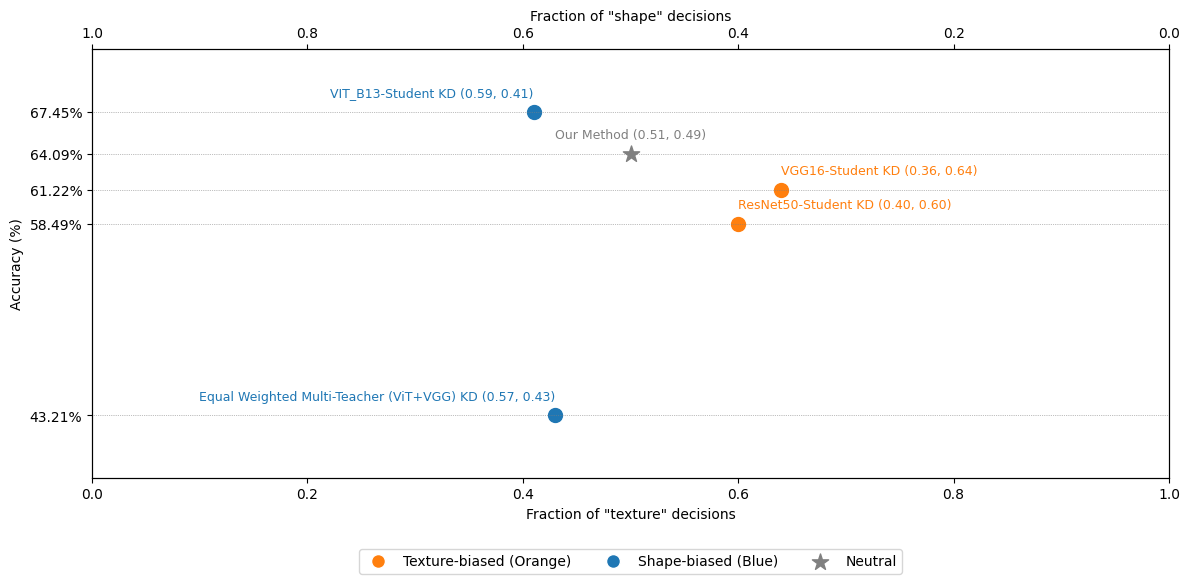}  
    \caption{Texture vs. Shape Decision Fractions for various employed methods as baselines \textit{(*Distillation was performed on limited epoch and smaller dataset due to compute and time constraints).}}
    \label{fig:model_analysis}
\end{figure}

In Figure~\ref{fig:model_analysis}, evaluation is conducted in terms of shape and texture biases alongside the accuracy of the model on the original dataset with a couple of baselines.\

In the first set of experiments, knowledge was distilled from a single teacher to a student model, we observe that teacher biases are consistently propagated to the student. In distillation from VGG-16 to a student model, the resulting student model exhibits an accuracy of \textbf{61.22\%}, with a texture bias of \textbf{0.64} and a shape bias of \textbf{0.36}. This indicates that despite the student being a ViT (Vision Transformer) model, which typically emphasizes shape cues, the texture bias of the VGG-16 teacher is still inherited. A similar trend is observed for ResNet-50, another CNN-based teacher model.

On distilling from a transformer-based teacher (ViT-B16), the student model inherits a shape bias of \textbf{0.59}, while texture of only \textbf{0.41}. We achieved the highest accuracy at \textbf{67.45\%} due to same architectures. These findings highlight the significant influence of teacher biases on student models during knowledge distillation.

The equal weighting method performed poorly, achieving an accuracy of only \textbf{43.21\%}. This highlights the importance of employing dynamically weighted teacher contributions during distillation. Equal weighting fails to leverage the complementary strengths of multiple teachers, resulting in poor performance.

Our proposed dynamically weighted knowledge distillation method achieves a notable balance between accuracy and bias generalization as show by the \textbf{star} in Figure~\ref{fig:model_analysis}. Although there is a slight trade-off in accuracy which is \textbf{64.09\%}. Our approach results in nearly equal biases, with a shape bias of \textbf{0.49} and a texture bias of \textbf{0.51}. This balance ensures that the distilled model is equally generalizable to both shape- and texture-based cues in out-of-distribution data.

\begin{table}[H]
\centering
\begin{adjustbox}{max width=\textwidth}
\begin{tabular}{|>{\columncolor[HTML]{C0C0C0}}l|>{\columncolor[HTML]{C4DDFF}}c|>{\columncolor[HTML]{FFCE93}}c|}
\hline
\textbf{Methods}    & \multicolumn{2}{|c|}{\textbf{ImageNet-1K}}         \\ \hline
                   & {Stylized Accuracy } & {Canny Edge Accuracy}          \\ \hline
{VGG-16 to Student KD}    & {29.14\%}       & {17.34\%}             \\ \hline
{ViT B-16 to Student KD}    & 21.89\%                &   30.20\%                   \\ \hline
{Equal Weighted Multi-Teacher KD}  & 11.43\%     & 7.06\%             \\ \hline
{Confidence Aware Multi-Teacher KD} &  \textbf{28.89\%}                &  \textbf{31.12\%}                      \\ \hline
\end{tabular}
\end{adjustbox}
\caption{OOD performance of Student Model post Knowledge Distillation.}
\label{tab:OOD_comparison}
\end{table}

As we can see in Table \ref{tab:OOD_comparison}, in all other methods, the student performed better towards a certain cue only, either stylized or Canny edge, depending on the teacher. In our method, it performed well on both the stylized and Canny edge datasets. Specifically, it achieved an accuracy of \textbf{28.89\%} on the stylized dataset and \textbf{31.12\%} on the Canny edge dataset. Although there is a very minute drop in accuracy on the stylized dataset as compared to the VGG16-student model, where it was \textbf{29.14\%}, we achieve a much more OOD generalizable student model, as it performs well on both datasets rather than just one of the cues.

\section{Discussions}

Our experiments reveal insights into the inductive biases of CNNs and Vision Transformers (ViTs) and the effects of knowledge distillation on student model. CNNs, like VGG-16, excel at handling texture-rich datasets, achieving the highest accuracy on standard datasets due to their ability to capture local texture features. However, their performance on Canny edge datasets is lower, highlighting a limitation in capturing global shape patterns.

ViTs, on the other hand, perform better on the Canny edge dataset, with ViT-Base achieving the highest accuracy of \textbf{55.83\% on ImageNet}, showcasing their strength in shape recognition. However, they struggle with stylized datasets, with ViT-Small achieving just 9.89\% on Animals-10, underlining their challenge in processing local texture cues.

Bias analysis confirmed that CNNs are predominantly texture-biased, while ViTs tend to be more shape-biased, for example ViT-Base, which showed a \textbf{64\% shape bias}. Surprisingly, ViT-Small exhibited a relatively balanced bias, though shape features still dominated.

In terms of knowledge distillation, we found that teacher model biases are transferred to student models. Distilling from texture-biased models like VGG-16 resulted in students inheriting texture bias, while distilling from shape-biased models like ViT-Base led to shape bias in students. Our dynamically weighted distillation method performed best, achieving a balance between accuracy (64.09\%) and bias generalization at \textbf{0.49 shape vs. 0.51 texture}, outperforming other methods like equal weighting, which performed poorly at \textbf{43.21\%} accuracy.

Overall, OOD generalization was improved with our approach, as the student model performed well on both stylized and Canny edge datasets, unlike other methods that excelled on just one cue. Although there was a slight drop in accuracy on stylized data compared to VGG16-student (29.14\%), our method achieved a more generalizable model across both datasets.

On the whole, CNNs and ViTs exhibit complementary biases, with CNNs favoring texture and ViTs favoring shape. By using dynamically weighted distillation, we effectively combine these strengths, resulting in a more robust and generalized model for diverse tasks.

\section{Conclusion}
In this study, we examined the inductive biases of CNNs and Vision Transformers and their impact on OOD generalization. Our results show that CNNs, like VGG-16 perform good in texture rich datasets, while ViTs perform better with shape based cues. We demonstrated that knowledge distillation effectively transfers these biases from teacher to student models, and our proposed dynamically weighted distillation method strikes a balance between accuracy and bias generalization. This approach outperformed traditional methods, leading to a student model that generalizes well across both stylized and Canny edge datasets, making it more apt for diverse tasks.

While our method achieves a balance between bias generalization and accuracy, there is a slight trade-off in the original dataset performance. This can be attributed to the nature of the heterogeneous setup and simple logit matching KD. We have observed how during distillation a teacher model can propagate its biases to the student model even when architectures differ; our method tackles the issue of mitigating these biases and produces a well-balanced student model.

Future work can focus on enhancing the dynamic weight allocation for teacher models and exploring hint-based distillation to improve accuracy and generalization. While hint-based distillation has previously been unsuccessful, if we succeed in aligning the representations of models like VGG and ViT-B16 into a common latent space, it could significantly improve accuracy and make our model more generalizable.

\textbf{Limitations}\\
While our method balances shape and texture biases, it does not outperform the Single-Teacher ViT in accuracy. This trade-off stems from challenges in heterogeneous architecture, where mismatched model capacities hinder effective learning. Discrepancies in the logits' scale and distribution due to different datasets can impair convergence. Additionally, the lack of feature-based distillation limits knowledge transfer, focusing only on final output probabilities, which may reduce accuracy on the original dataset, advocating for exploration in feature based and structural distillation in a heterogeneous environment.

\bibliographystyle{plainnat}
\bibliography{ref}

\appendix
\section{Model Architectures and Hyperparameters}
\label{sec:appendix}

This appendix reports the exact architectures, preprocessing, and training hyperparameters used in our experiments, extracted directly from our implementation, to support reproducibility.

\subsection{Model Architectures}

\begin{table}[H]
\centering
\begin{adjustbox}{max width=\textwidth}
\begin{tabular}{@{}llll@{}}
\toprule
\textbf{Role} & \textbf{Model} & \textbf{Source} & \textbf{Trainable parameters} \\
\midrule
Teacher A (texture) & VGG-16 & torchvision, ImageNet-1K pretrained & classifier head only (backbone frozen) \\
Teacher B (shape) & ViT-B/16 & torchvision, ImageNet-1K pretrained & classification head only (backbone frozen) \\
Student & ViT-Small/16 (\texttt{vit\_small\_patch16\_224}) & \texttt{timm} & fully trainable \\
\bottomrule
\end{tabular}
\end{adjustbox}
\caption{Model sources and which parameters were updated during training. All models operate on $224\times224$ inputs.}
\label{tab:appendix-models}
\end{table}

\subsection{Data Preprocessing}
All images were resized to $224\times224$ and normalized with standard ImageNet statistics ($\mu = [0.485, 0.456, 0.406]$, $\sigma = [0.229, 0.224, 0.225]$). No additional data augmentation (random crop, flip, color jitter) was applied. Train/validation splits used an 80/20 ratio with a fixed random seed of 42.

\subsection{Training Hyperparameters}

\begin{table}[H]
\centering
\begin{adjustbox}{max width=\textwidth}
\begin{tabular}{@{}llllll@{}}
\toprule
\textbf{Stage} & \textbf{Epochs} & \textbf{Batch size} & \textbf{Optimizer} & \textbf{LR} & \textbf{Other} \\
\midrule
Teacher fine-tuning (VGG-16, ViT-B/16) & 10 & 128 & Adam & $1\times10^{-4}$ & classifier head only \\
Student warm-up (CE only) & 2 & 128 & Adam & $1\times10^{-4}$ & full fine-tuning \\
Confidence-aware multi-teacher KD & 8 & 32 & AdamW ($\lambda_{wd}=0.01$) & $1\times10^{-3}$ & cosine annealing, grad.\ clip (max norm 1.0) \\
\bottomrule
\end{tabular}
\end{adjustbox}
\caption{Training hyperparameters by stage. The student is first warmed up with cross-entropy only, then trained with our full confidence-aware KD objective (Eq.~\ref{eq:total-loss}).}
\label{tab:appendix-hparams}
\end{table}

For the confidence-aware KD stage specifically: student and teacher temperatures $T_s = T_t = 2.0$ (Eq.~\ref{eq:distill}), loss balance $\alpha = 0.5$ (Eq.~\ref{eq:total-loss}), confidence-weighting softmax temperature $\beta = 1.5$ (Eq.~\ref{eq:weight-norm}), and numerical stabilizers $\epsilon = 10^{-10}$ (entropy) and $\epsilon = 10^{-6}$ (z-score standard deviation).


\end{document}